\documentclass[runningheads]{llncs}
\usepackage{graphicx}
\usepackage{comment}
\usepackage{amsmath,amssymb} 
\usepackage{color}
\usepackage{multirow}
\usepackage{booktabs}
\usepackage{array}

\makeatletter
\def\thickhline{%
	\noalign{\ifnum0=`}\fi\hrule \@height \thickarrayrulewidth \futurelet
	\reserved@a\@xthickhline}
\def\@xthickhline{\ifx\reserved@a\thickhline
	\vskip\doublerulesep
	\vskip-\thickarrayrulewidth
	\fi
	\ifnum0=`{\fi}}
\makeatother

\newlength{\thickarrayrulewidth}
\begin{document}
\pagestyle{headings}
\mainmatter
\def\ECCVSubNumber{990}  

\title{Circumventing Outliers of AutoAugment\\with Knowledge Distillation} 

\titlerunning{ECCV-20 submission ID \ECCVSubNumber} 
\authorrunning{ECCV-20 submission ID \ECCVSubNumber} 
\author{Anonymous ECCV submission}
\institute{Paper ID \ECCVSubNumber}

\titlerunning{Circumventing Outliers of AutoAugment with Knowledge Distillation}

\author{\!Longhui Wei\textsuperscript{1}, An Xiao\textsuperscript{1}\thanks{The first two authors contributed equally to this work.}, Lingxi Xie\textsuperscript{1}, Xin Chen\textsuperscript{1,2}, Xiaopeng Zhang\textsuperscript{1}, Qi Tian\textsuperscript{1}\!}
\authorrunning{L. Wei, A. Xiao, et al.}
%
\institute{\textsuperscript{1}Huawei Inc.,\quad\textsuperscript{2}Tongji University\\
\email{\{weilonghui1,xiaoan1\}@huawei.com}, \email{198808xc@gmail.com}\\
\email{1410452@tongji.edu.cn}, \email{zxphistory@gmail.com}, \email{tian.qi1@huawei.com}}
\maketitle

\begin{abstract}
AutoAugment has been a powerful algorithm that improves the accuracy of many vision tasks, yet it is sensitive to the operator space as well as hyper-parameters, and an improper setting may degenerate network optimization. This paper delves deep into the working mechanism, and reveals that AutoAugment may remove part of discriminative information from the training image and so insisting on the ground-truth label is no longer the best option. To relieve the inaccuracy of supervision, we make use of knowledge distillation that refers to the output of a teacher model to guide network training. Experiments are performed in standard image classification benchmarks, and demonstrate the effectiveness of our approach in suppressing noise of data augmentation and stabilizing training. Upon the cooperation of knowledge distillation and AutoAugment, we claim the \textbf{new state-of-the-art} on ImageNet classification with a top-$1$ accuracy of $\mathbf{85.8\%}$.
\end{abstract}

\section{Introduction}
\label{introduction}

Automated machine learning (AutoML) has been attracting increasing attentions in recent years. In standard image classification tasks, there are mainly two categories of AutoML techniques, namely, neural architecture search (NAS) and hyper-parameter optimization (HPO), both of which focus on the possibility of using automatically learned strategies to replace human expertise. AutoAugment~\cite{cubuk2019autoaugment} belongs to the latter, which goes one step beyond conventional data augmentation techniques (\textit{e.g.}, horizontal flipping, image rescaling \& cropping, color jittering, \textit{etc.}) and tries to combine them towards generating more training data without labeling new images. It has achieved consistent accuracy gain in image classification~\cite{cubuk2019autoaugment}, object detection~\cite{ghiasi2019fpn}, \textit{etc.}, and meanwhile efficient variants of AutoAugment have been proposed to reduce the computational burden in the search stage~\cite{ho2019population,lim2019fast,hataya2019faster}.

\begin{figure}[!t]
\centering
\includegraphics[width=12cm,height=4cm]{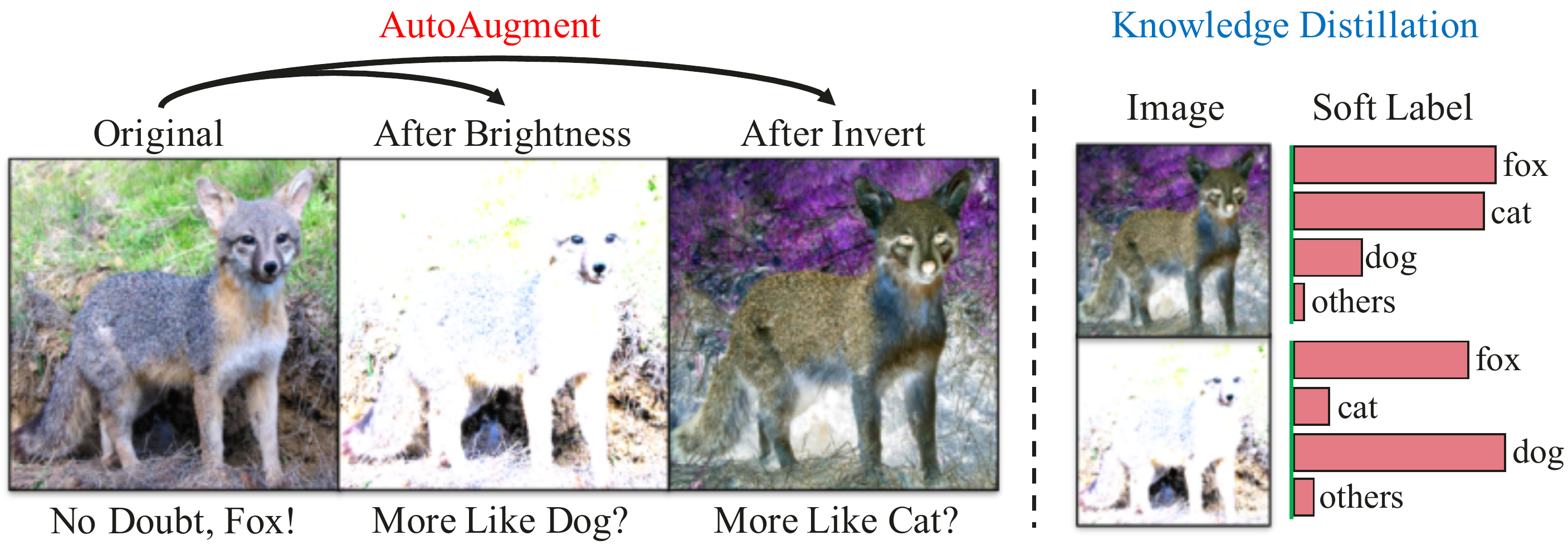}
\caption{\textbf{Left}: an image and its augmented copies generated by AutoAugment. The original image is clean and there is no doubt to use the ground-truth label, while the augmented counterparts look more like other classes which the annotation is not aware of. This phenomenon is called \textbf{augment ambiguity}. \textbf{Right}: We leverage the idea of knowledge distillation to provide softened signals to avoid ambiguity.}
\label{fig:introduction}
\end{figure}

Despite their ability in improving recognition accuracy, we note that AutoAu-\\gment-based methods often require the search space to be well-designed. Without careful control (\textit{e.g.}, in an expanded search space or with an increased distortion magnitude), these methods are not guaranteed to perform well -- as we shall see in Section~\ref{approach:auto_augment}, an improper hyper-parameter may deteriorate the optimization process, resulting in even lower accuracy compared to the baseline. This puts forward a hard choice between more information (seeing a wider range of augmented images) and safer supervision (restricting the augmented image within a relatively small neighborhood around the clean image), which downgrades the upper-bound of AutoAugment-based methods.

In this paper, we investigate the reason of this contradictory. We find that when heavy data augmentation is added to the training image, it is probable that part of its semantic information is removed. An example of changing image brightness is shown in Figure~\ref{fig:introduction}, and other transformation such as image translation and shearing can also incur information loss and make the image class unrecognizable (refer to Figure~\ref{fig:space}). We name this phenomenon \textbf{augment ambiguity}. In such contaminated training images, insisting on the ground-truth label is no longer the best choice, as the inconsistency between input and supervision can confuse the network. Intuitively, complementary information that relates the augmented image to similar classes may serve as a better source of supervision.

Motivated by the above, we leverage the idea of knowledge distillation which uses a standalone model (often referred to as the teacher) to guide the target network (often referred to as the student). For each augmented image, the student receives supervision from both the ground-truth label and the teacher signal, and in case that part of semantic information is removed, the teacher model is expected to provide softened labels to avoid confusion. The extra loss between teacher and student is measured by the KL-divergence between the score distributions of their top-ranked classes, and the number of involved classes is positively correlated to the magnitude of augmentation, since a larger magnitude often eliminates more semantics and causes smoother score distributions.

The main contribution of this paper is to reveal that \textit{knowledge distillation is a natural complement to uncontrolled data augmentation, such as AutoAugment and its variants}. The effectiveness of our approach is verified in the space of AutoAugment~\cite{cubuk2019autoaugment} as well as that of RandAugment~\cite{cubuk2019randaugment} with different strengths of transformations. Knowledge distillation brings consistent accuracy gain to recognition, in particular when the distortion magnitude becomes larger. Experiments are performed on standard image classification benchmarks, namely, CIFAR-10/100 and ImageNet. On CIFAR-100, with a strong baseline of PyramidNet~\cite{han2017deep} and ShakeDrop~\cite{yamada2019shakedrop} regularization, we achieve a test error of $10.6\%$, outperforming all competitors with similar training costs. On ImageNet, in the RandAugment space, we boost the top-1 accuracy of EfficientNet-B7~\cite{tan2019efficientnet} from $84.9\%$ to $85.5\%$, with a significant improvement of $0.6\%$. Note that without knowledge distillation, RandAugment with a large distortion magnitude may suffer unstable training. Moreover, on top of EfficientNet-B8~\cite{tan2019efficientnet}, we set a \textbf{new record} on ImageNet classification (without extra training data) by claiming a top-$1$ accuracy of $\mathbf{85.8\%}$, which surpasses the previous best by a non-trivial margin of $0.3\%$.

The remaining part of this paper is organized as follows. Section~\ref{related_work} briefly reviews related work, and Section~\ref{approach} elaborates our motivation and approach. After experiments are shown in Section~\ref{experiments}, we conclude this work in Section~\ref{conclusions}.

\section{Related Work}
\label{related_work}

Deep learning~\cite{lecun2015deep}, in particular training deep neural networks, has been the standard methodology in computer vision. Modern neural networks, either manually-designed~\cite{krizhevsky2012imagenet,simonyan2015very,szegedy2015going,he2016deep,huang2017densely} or automatically searched~\cite{zoph2017neural,real2017large,zoph2018learning,liu2018progressive,tan2019mnasnet,tan2019efficientnet}, often contain a very large number of trainable parameters and thus raise the challenge of collecting more labeled data to avoid over-fitting. Data augmentation is a standard strategy to generate training data without additional labeling costs. Popular options of transformation include horizontal flipping, color/contrast jittering, image rotation/shearing, \textit{etc}., each of which slightly alters the geometry and/or pattern of an image but keeps its semantics (\textit{e.g.}, the class label) unchanged. Data augmentation has been verified successful in a wide range of visual recognition tasks, including image classification~\cite{devries2017improved,zhang2018mixup,yun2019cutmix}, object detection~\cite{singh2017hide}, semantic segmentation~\cite{fang2019instaboost}, person re-identification~\cite{zhong2017random}, \textit{etc}. Researchers have also discussed the connection between data augmentation and network regularization~\cite{srivastava2014dropout,gastaldi2017shake,yamada2019shakedrop} methods.

With the rapid development of automated machine learning (AutoML)~\cite{thornton2013auto}, researchers proposed to learn data augmentation strategies in a large search space to replace the conventional hand-designed augmentation policies. AutoAugment~\cite{cubuk2019autoaugment} is one of the early efforts that works on this direction. It first designed a search space with a number of transformations and then applied reinforcement learning to search for powerful combinations of the transformations to arrive at a high validation accuracy. To alleviate the heavy computational costs in the search stage, researchers designed a few efficient variants of AutoAugment. 
Fast AutoAugment~\cite{lim2019fast} moved the costly search stage from training to evaluation through bayesian optimization, and population-based augmentation~\cite{ho2019population} applied evolutionary algorithms to generate policy schedule by only a single run of 16 child models. Online hyper-parameter learning~\cite{lin2019online} combined the search stage and the network training process, and faster AutoAugment~\cite{hataya2019faster} formulated the search process into a differentiable function, following the recently emerging weight-sharing NAS approaches~\cite{brock2018smash,pham2018efficient,liu2019darts}. Meanwhile, some properties of AutoAugment have been investigated, such as whether aggressive transformations need to be considered~\cite{ho2019population} and how transformations of enriched knowledge are effectively chosen~\cite{zhang2020adversarial}. Recently, RandAugment~\cite{cubuk2019randaugment} shared another opinion that the search space itself may have contributed most: based on a well-designed set of transformations, a random policy of augmentation works sufficiently well.

Knowledge distillation was first introduced as an approach to assist network optimization~\cite{hinton2015distilling}. The goal is to improve the performance of a target network (often referred to as the student model) using two sources of supervision, one from the ground-truth label, and the other from the output signal of a pre-trained network (often referred to as the teacher model). Beyond its wide application on model compression (large teacher, small student~\cite{romero2014fitnets,hinton2015distilling}) and model initialization (small teacher, large student~\cite{simonyan2015very,chen2016net2net}), researchers later proposed to use it for standard network training, with the teacher and student models sharing the same network architecture~\cite{furlanello2018born,yang2019training,yang2019snapshot}, and sometimes under the setting of semi-supervised learning~\cite{tarvainen2017mean,zhang2018deep}. There have been discussions on the working mechanism of knowledge distillation, and researchers advocated for the so-called `dark knowledge'~\cite{hinton2015distilling} being some kind of auxiliary supervision, obtained from the pre-trained model~\cite{yang2019training,bagherinezhad2018label} and thus different from the softened signals based on label smoothing~\cite{szegedy2016rethinking,pereyra2017regularizing}.

In this paper, we build the connection between knowledge distillation and AutoAugment by showing that the former is a natural complement to the latter, which filters out noises introduced by overly aggressive transformations.

\section{Our Approach}
\label{approach}

\subsection{Preliminaries: Data Augmentation with AutoML}
\label{approach:preliminaries}

Let ${\mathcal{D}}={\left\{\left(\mathbf{x}_n,y_n\right)\right\}_{n=1}^N}$ be a labeled image classification dataset with $N$ samples, in which $\mathbf{x}_n$ denotes the raw pixels and $y_n$ denotes the annotated label within the range of $\left\{1,2,\ldots,C\right\}$, $C$ is the number of classes. $\mathbf{y}_n$ is the vectorized version of $y_n$ with the dimension corresponding to the true class assigned a value of $1$ and all others being $0$. The goal is to train a deep network, ${\mathbb{M}}:{\mathbf{y}_n}={\mathbf{f}\!\left(\mathbf{x}_n;\boldsymbol{\theta}\right)}$, where $\boldsymbol{\theta}$ denotes the trainable parameters. The dimensionality of $\boldsymbol{\theta}$ is often very large, \textit{e.g.}, tens of millions, exceeding the size of dataset, $N$, in most of cases. Therefore, network training is often a ill-posed optimization problem and incurs over-fitting without well-designed training strategies.

The goal of data augmentation is to enlarge the set of training images without actually collecting and annotating more data. It starts with defining a transformation function, ${\mathbf{x}_n^{\boldsymbol{\tau}}}\doteq{\mathbf{g}\!\left(\mathbf{x}_n,\boldsymbol{\tau}\right)}$, in which ${\boldsymbol{\tau}}\sim{\mathcal{T}}$ is a multi-dimensional vector parameterizing how the transformations are performed. Note that each dimension of $\boldsymbol{\tau}$ can take either discrete (\textit{e.g.}, whether the image is horizontally flipped) or continuous (\textit{e.g.}, the image is rotated for a specific angle), and different transformations can be applied to an image towards richer combinations. The idea of AutoAugment~\cite{cubuk2019autoaugment} is to optimize the distribution, $\mathcal{T}$, so that the model optimized on the augmented training set achieves satisfying performance on the validation set:
\begin{eqnarray}
\label{eqn:auto_augment}
{\mathcal{T}^\star}&=&{\arg\min_\mathcal{T}\mathcal{L}\!\left(\mathbf{g}\!\left(\mathbf{x}_n;\boldsymbol{\tau}\sim\mathcal{T}\right),\mathbf{y}_n;\boldsymbol{\theta}_\mathcal{T}^\star\mid\left(\mathbf{x}_n,\mathbf{y}_n\right)\sim\mathcal{D}_\mathrm{val}\right)},\\
\nonumber
\mathrm{in\ which\quad\quad}{\boldsymbol{\theta}_\mathcal{T}^\star}&=&{\arg\min_{\boldsymbol{\theta}}\mathcal{L}\!\left(\mathbf{g}\!\left(\mathbf{x}_n;\boldsymbol{\tau}\sim\mathcal{T}\right),\mathbf{y}_n;\boldsymbol{\theta}\mid\left(\mathbf{x}_n,\mathbf{y}_n\right)\sim\mathcal{D}_\mathrm{train}\right)}.
\end{eqnarray}
Here, $\mathcal{D}_\mathrm{train}$ and $\mathcal{D}_\mathrm{val}$ are two subsets of $\mathcal{D}$, used for training and validating the quality of $\mathcal{T}$, respectively. The loss function follows any conventions, \textit{e.g.}, the cross-entropy form:
\begin{equation}
\label{eqn:cross_entropy}
{\mathcal{L}\!\left(\mathbf{x}_n^{\boldsymbol{\tau}},\mathbf{y}_n;\boldsymbol{\theta}\right)}={\mathbf{y}_n^\top\cdot\ln\mathbf{f}\!\left(\mathbf{x}_n^{\boldsymbol{\tau}};\boldsymbol{\theta}\right)},
\end{equation}
Eqn~\eqref{eqn:auto_augment} is a two-stage optimization problem, for which existing approaches either applied reinforcement learning~\cite{cubuk2019autoaugment,lim2019fast,ho2019population} or weight-sharing methods~\cite{hataya2019faster} which are often more efficient.

We follow the convention to assign each dimension in $\boldsymbol{\tau}$ to be an individual transformation, with the complete list shown below:
\begin{table}[!h]
\small
\centering
\vspace{-0.5cm}
\begin{tabular}{llll}
$\bullet\;$\texttt{invert} $\qquad$    & $\bullet\;$ \texttt{autoContrast} $\qquad$ &
$\bullet\;$ \texttt{equalize} $\qquad$  & $\bullet\;$ \texttt{rotate} $\qquad$   \\
$\bullet\;$\texttt{solarize} $\qquad$    & $\bullet\;$ \texttt{color} $\qquad$        &
$\bullet\;$ \texttt{posterize} $\qquad$ & $\bullet\;$ \texttt{contrast} $\qquad$ \\
$\bullet\;$\texttt{brightness} $\qquad$  & $\bullet\;$ \texttt{sharpness} $\qquad$    &
$\bullet\;$ \texttt{shear-x} $\qquad$   & $\bullet\;$ \texttt{shear-y} $\qquad$  \\
$\bullet\;$\texttt{translate-x} $\qquad$ & $\bullet\;$ \texttt{translate-y} $\qquad$ \\
\end{tabular}
\vspace{-0.6cm}
\end{table}

\noindent
Therefore, $\boldsymbol{\tau}$ is a 14-dimensional vector and each dimension of $\boldsymbol{\tau}$ represents the magnitude of the corresponding transformation. For example, the fourth dimension of $\boldsymbol{\tau}$ represents the magnitude of  \texttt{rotate} transformation, and a value of zero indicates the corresponding transformation being switched off. Each time a transformation is sampled from the distribution, ${\boldsymbol{\tau}}\sim{\mathcal{T}}$, at most two dimensions in it are set to be non-zero, and each selected transformation is assigned a probability that it is applied after each training image is sampled online.

\subsection{AutoAugment Introduces Noisy Training Images}
\label{approach:auto_augment}

\begin{figure}[!t]
\centering
\includegraphics[width=12cm,height=4cm]{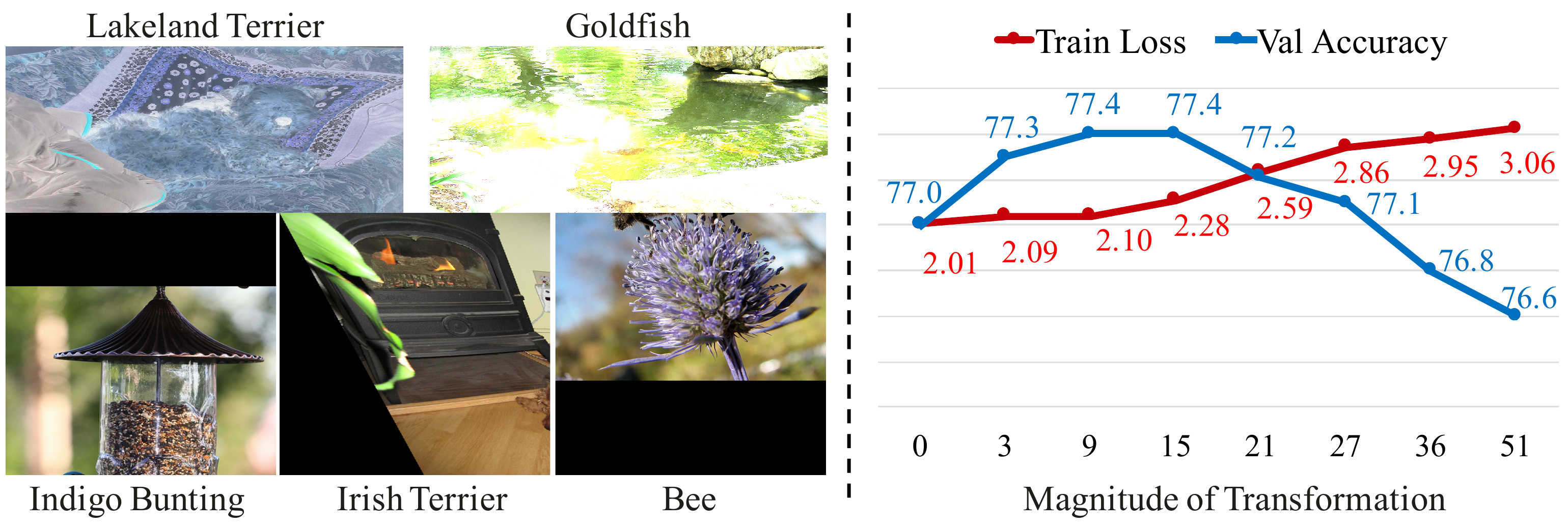}
\caption{\textbf{Left}: AutoAugment can generate meaningless training images but still assigns deterministic class labels to them. \textbf{Right}: The results of EfficientNet-B0 with different magnitudes of transformation on ImageNet. The training difficulty increases gradually with enlarging the magnitude of transformation, while the validation accuracy rises initially but drops at last. This phenomenon reveals the model starts from over-fitting to under-fitting.}
\label{fig:motivation}
\end{figure}

AutoAugment makes it possible to generate infinitely many images which do not exist in the original training set. On the upside, this reduces the risk of over-fitting during the training process; on the downside, it can introduce a considerable amount of outliers to the training process. Typical examples are shown in Figure~\ref{fig:motivation}. When an image with its upper part occupied by main content (\textit{e.g.}, \textit{bee}) is sampled, the transformation of \texttt{translate-y} (shifting the image along the vertical direction) suffers risk of removing all discriminative contents within it outside the visible area, and thus the augmented image becomes meaningless in semantics. Nonetheless, the training process is not always aware of such noises and still uses the ground-truth signal, a one-hot vector, to supervise and thus confuse the deep network.

In Figure~\ref{fig:motivation}, we also show how the training loss and validation accuracy curves change along with the magnitude of transformation. When the magnitude is $0$ (\textit{i.e.}, no augmentation is used), it is easy for the network to fit the training set and thus the training loss quickly drops, but the validation accuracy remains low which indicates over-fitting. With a relatively low magnitude of augmentation, the training loss increases gradually meanwhile the validation accuracy arrives at a higher plateau, \textit{i.e.}, over-fitting is alleviated. However, if the magnitude of augmentation continues growing, it becomes more and more difficult to fit the training set, \textit{i.e.}, the model suffers under-fitting. In particular, when the magnitude is set to be $36$, the noisy data introduced to the training set is sufficiently high to bias the model training,~\textit{i.e.}, the results is lower than the baseline without AutoAugment.

From the above analysis, we realize that AutoAugment is indeed balancing between richer training data and heavier noises. Researchers provided comments from two aspects: some of them argued that the transformation strategies may have been overly aggressive and need to be controlled~\cite{ho2019population}, while some others advocated for the benefit of exploring aggressive transformations so that richer information is integrated into the trained model~\cite{zhang2020adversarial}. We deal with this issue from a new perspective. We believe that aggressive transformations are useful to training, yet treating all augmented images just like they are clean (non-augmented) samples is not the optimal choice. Moreover, the same transformations operated on different images will cause different results,~\textit{i.e.}, some generated images can enrich the diversity of training set but the others are biased. Therefore, we treat every image differently for preserving the richer information but filtering out the noises. 

\subsection{Circumventing Outliers with Knowledge Distillation}
\label{approach:kl_divergence}

Our idea is very simple. For a training image generated by AutoAugment, $\mathbf{g}\!\left(\mathbf{x}_n;\boldsymbol{\tau}\right)$, we provide two-source supervision signals to guide network optimization. The first one remains the same as the original training process, with the standard cross-entropy loss computed based on the ground-truth class, $\mathbf{y}_n$. The second one comes from a pre-trained model which provides an individual judgment of $\mathbf{g}\!\left(\mathbf{x}_n;\boldsymbol{\tau}\right)$, \textit{i.e.}, whether it contains sufficient semantics for classification. Let $\mathbb{M}^\mathrm{T}$ and $\mathbb{M}^\mathrm{S}$ denote the pre-trained (teacher) and target (student) model, where the superscripts of $\mathrm{T}$ and $\mathrm{S}$ represent `teacher' and `student', respectively, and thus Eqn~\eqref{eqn:cross_entropy} is upgraded to be:
\begin{equation}
\label{eqn:knowledge_distillation}
{\mathcal{L}^\mathrm{KD}\!\left(\mathbf{x}_n^{\boldsymbol{\tau}},\mathbf{y}_n;\boldsymbol{\theta}^\mathrm{S}\right)}={\mathbf{y}_n^\top\cdot\ln\mathbf{f}^\mathrm{S}\!\left(\mathbf{x}_n^{\boldsymbol{\tau}};\boldsymbol{\theta}^\mathrm{S}\right)+\lambda\cdot\mathrm{KL}\!\left[\mathbf{f}^\mathrm{S}\!\left(\mathbf{x}_n^{\boldsymbol{\tau}};\boldsymbol{\theta}^\mathrm{S}\right)\|\mathbf{f}^\mathrm{T}\!\left(\mathbf{x}_n^{\boldsymbol{\tau}};\boldsymbol{\theta}^\mathrm{T}\right)\right]},
\end{equation}
where $\lambda$ is the balancing coefficient, and we have followed the convention to use the KL-divergence to compute the distance between teacher and student outputs, two probabilistic distributions over all classes.

Intuitively, when the semantic information of an image is damaged by data augmentation, the teacher model that is `unaware' of augmentation should produce less confident probabilistic outputs, \textit{e.g.}, if an original image, $\mathbf{x}_n$, contains a specific kind of \textit{bird} and some parts of the \textit{bird} is missing or contaminated by augmentation, $\boldsymbol{\tau}$, then we expect the probabilistic scores of the augmented image, $\mathbf{x}_n^{\boldsymbol{\tau}}$, to be distributed over a few classes with close relationship to the true one. We introduce a hyper-parameter, $K$, and consider the $K$ classes with the highest scores in $\mathbf{f}^\mathrm{T}\!\left(\mathbf{x}_n^{\boldsymbol{\tau}};\boldsymbol{\theta}^\mathrm{T}\right)$, forming a set denoted by $\mathcal{C}_K\!\left(\mathbf{x}_n^{\boldsymbol{\tau}};\boldsymbol{\theta}^\mathrm{T}\right)$. Most often, we have ${K}\ll{C}$, and the choice of $K$ will be discussed empirically in the experimental section. The KL-divergence between $\mathbf{f}^\mathrm{T}\!\left(\mathbf{x}_n^{\boldsymbol{\tau}};\boldsymbol{\theta}^\mathrm{T}\right)$ and $\mathbf{f}^\mathrm{S}\!\left(\mathbf{x}_n^{\boldsymbol{\tau}};\boldsymbol{\theta}^\mathrm{S}\right)$ is thus modified as:
\begin{equation}
\label{eqn:kl}
{\mathrm{KL}\!\left[\mathbf{f}^\mathrm{S}\!\left(\mathbf{x}_n^{\boldsymbol{\tau}};\boldsymbol{\theta}^\mathrm{S}\right)\|\mathbf{f}^\mathrm{T}\!\left(\mathbf{x}_n^{\boldsymbol{\tau}};\boldsymbol{\theta}^\mathrm{T}\right)\right]}={\sum_{c\in\mathcal{C}_K\!\left(\mathbf{x}_n^{\boldsymbol{\tau}};\boldsymbol{\theta}^\mathrm{T}\right)}f_c^\mathrm{T}\!\left(\mathbf{x}_n^{\boldsymbol{\tau}};\boldsymbol{\theta}^\mathrm{T}\right)\cdot\ln\frac{f_c^\mathrm{S}\!\left(\mathbf{x}_n^{\boldsymbol{\tau}};\boldsymbol{\theta}^\mathrm{S}\right)}{f_c^\mathrm{T}\!\left(\mathbf{x}_n^{\boldsymbol{\tau}};\boldsymbol{\theta}^\mathrm{T}\right)}},
\end{equation}
where $f_c$ denotes the $c$-th dimension of $\mathbf{f}$.

\subsection{Discussions and Relationship with Prior Work}
\label{approach:discussions}

A few prior work~\cite{bagherinezhad2018label,yang2019training} studied how knowledge distillation works in the scenarios that teacher and student models have the same capacity. They argued that the teacher model should be strong enough so as not to provide low-quality supervision to the student model. However, this work provides a novel usage of the teacher signal: suppressing noises introduced by data augmentation. From this perspective, the teacher model can be considerably weaker than the student model but still contribute to recognition accuracy. Experimental results on CIFAR-100 (setting and details are provided in Section~\ref{experiments:cifar}) show that a pre-trained Wide-ResNet-28-10~\cite{Zagoruyko2016} with AutoAugment (test set error rate of $17.1\%$) can reduce the test set error rate of a Shake-Shake (26 2x96D)~\cite{gastaldi2017shake} trained with AutoAugment from $14.3\%$ to $13.8\%$.

We noticed prior work~\cite{he2019data} argued that data augmentation may introduce uncertainty to the network training process because the training data distribution is changed, and proposed to switch off data augmentation at the end of the training stage to alleviate the empirical risk of optimization. Our method provides an alternative perspective that the risk is likely to be caused by the noises of data augmentation and thus can be reduced by knowledge distillation. Moreover, the hyper-parameters in~\cite{he2019data} (\textit{e.g.}, when to switch off data augmentation) is difficult to tune. In training Wide-ResNet-28-10~\cite{Zagoruyko2016} with AutoAugment on CIFAR-100, we follow the original paper to prevent data augmentation by adding $50$ epochs to train the clean images only, but the baseline error rate ($17.1\%$) is only reduced to $16.8\%$. In comparison, when knowledge distillation is added to these $50$ epochs, the error rate is significantly reduced to $16.2\%$.

This work is also related to prior efforts that applied self-training to semi-supervised learning, \textit{i.e.}, only a small portion of training data is labeled~\cite{tarvainen2017mean,laine2017temporal,xie2019self}. These methods often started with training a model on the labeled part, then used this model to `guess' a pseudo label for each of the unlabeled samples, and finally updated the model using all data with either ground-truth or pseudo labels. This paper verifies the effectiveness of knowledge distillation in the fully-supervised setting in which augmented data can be noisy. Therefore, we draw the connection between exploring unseen data (data augmentation) and exploiting unlabeled data (semi-supervised learning), and reveal the potential of integrating AutoAugment and/or other hyper-parameter optimization methods to assist and improve semi-supervised learning.

\section{Experiments}
\label{experiments}

\subsection{On the CIFAR-10/100 Datasets}
\label{experiments:cifar}

\noindent$\bullet$\quad\textbf{Dataset and Settings}

CIFAR-10 and CIFAR-100~\cite{krizhevsky2009learning} contain tiny images with a resolution of $32\times32$. Both CIFAR-10 and CIFAR-100 have $50\mathrm{K}$ training and $10\mathrm{K}$ testing images, uniformly distributed over $10$ or $100$ classes. They are two commonly used datasets for validating the basic properties of learning algorithms.

Following the convention~\cite{cubuk2019autoaugment,cubuk2019randaugment}, we train three types of networks, namely, wide ResNet (Wide-ResNet-28-10)~\cite{Zagoruyko2016}, Shake-Shake (three variants with different feature dimensions)~\cite{gastaldi2017shake}, and PyramidNet~\cite{han2017deep} with ShakeDrop regularization~\cite{yamada2019shakedrop}.

\vspace{0.2cm}
\noindent$\bullet$\quad\textbf{Knowledge Distillation Stabilizes AutoAugment}

The core idea of our approach is to utilize knowledge distillation to restrain noises generated by severe transformations. This is expected to stabilize the training process of AutoAugment. To verify this, we start with training Wide-ResNet-28-10 on CIFAR-100. Note that the original augmentation space of AutoAugment involves two major kinds of transformations, namely, geometric or color-based transformations, on which AutoAugment as well as its variants limited the distortion magnitude of each transformation in a relatively small range so that the augmented images are mostly safe, \textit{i.e.}, semantic information is largely preserved. In order to enhance the benefit brought by suppressing noises of aggressive augmentations, we design a new augment space in which the restriction in distortion magnitude is much weaker. To guarantee that large magnitudes lead to complete damage of semantic information, we only preserve a subset of geometric transformations (\texttt{shear-x}, \texttt{shear-y}, \texttt{translate-x}, \texttt{translate-y}) as well as \texttt{cutout}, and set $10$ levels of distortion, so that ${M}={0}$ implies no augment, and ${M}={10}$ of any transformation destroys the entire image. Note that the range of $M$ here is specifically designed for the modified augment space, which is incomparable with the original definition of $M$ in RandAugment (experimented in Section~\ref{experiments:imagenet}). Regarding knowledge distillation, we set ${K}={3}$ (computing KL-divergence between the distributions of top-$3$ classes, determined by the teacher model) for CIFAR-10 and ${K}={5}$ for CIFAR-100. The balancing coefficient, $\lambda$, is set to be $1.0$.

\begin{figure}[!t]
\centering
\includegraphics[width=1\linewidth]{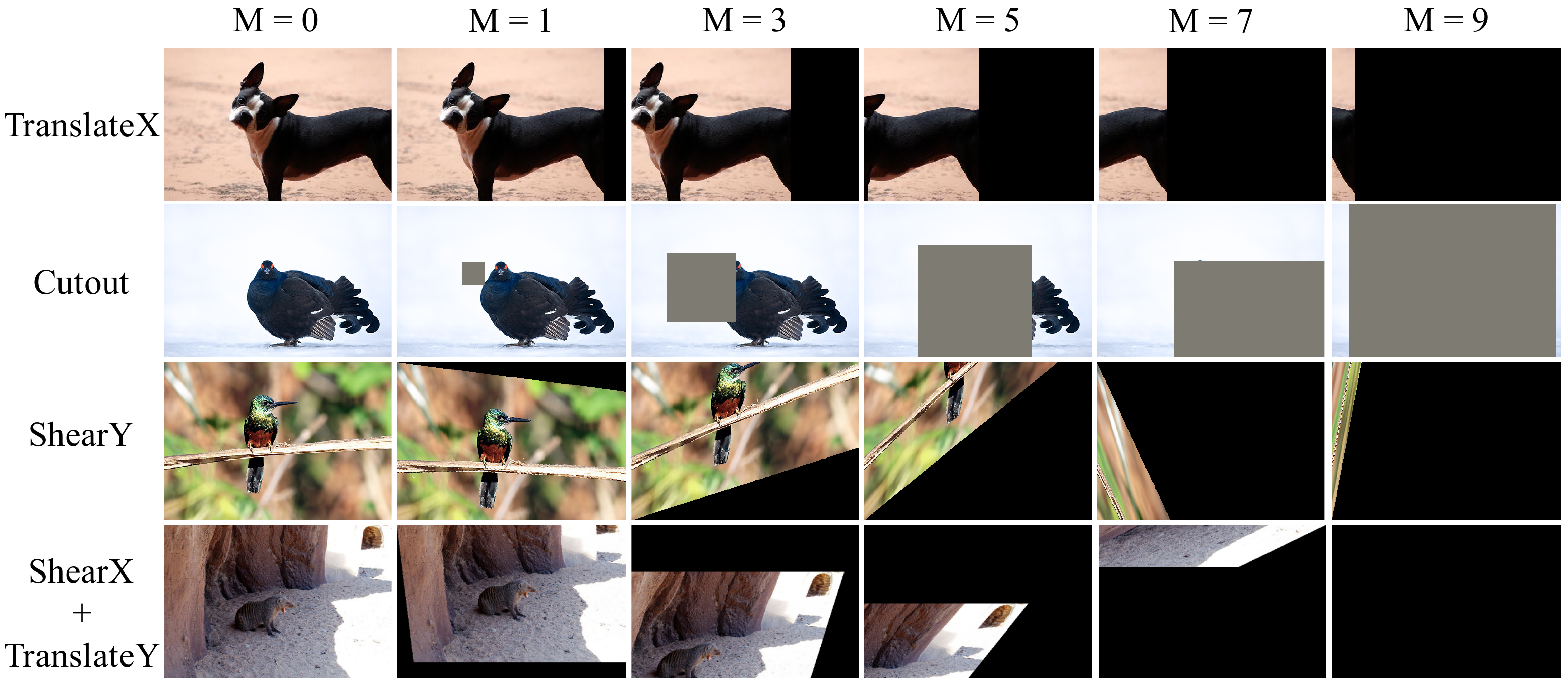}
\caption{Examples of transformations involved in our self-designed augment space. The distortion magnitude, $M$, is divided into $10$ levels. The deformation introduced by transformations increases along with the magnitude. First three rows are examples of the deformation produced by each type of transformation with different magnitudes. The last row represents applying two consecutive transformations on a single image, which is the real case in our training scenario.}
\label{fig:space}
\end{figure}

\begin{table}[!t]
\centering
\setlength{\abovecaptionskip}{10pt}%
\begin{tabular}{p{2.5cm}<{\centering}|p{1cm}<{\centering}p{1cm}<{\centering}p{1cm}<{\centering}p{1cm}<{\centering}p{1cm}<{\centering}p{1cm}<{\centering}p{1cm}<{\centering}p{1cm}<{\centering}}
\thickhline
\multirow{2}{2.5cm}{\centering \textbf{Model}}&\multicolumn{8}{c}{\textbf{Distortion Magnitude, $M$}}\\
\cline{2-9}
&0 & 1 & 2 & 3 & 4 & 5 & 6 & 7  \\
\hline
RA & 18.4 & 19.5 & 20.7 & 22.4 & 25.7 & 31.6 & 40.3 & 55.1 \\
RA+KD & 18.0 & 17.6 & 18.5 & 19.9 & 21.9 & 27.0 & 34.9 & 48.0 \\
\hline
Gain & +0.4 & +1.9 & +2.2 & +2.5 & +3.8 & +4.6 & +5.4 & +7.1 \\
\thickhline
\end{tabular}
\caption{Comparison between RandAugment with or without knowledege distillation in our self-designed augment space on CIFAR-100 based on Wide-ResNet-28-10. All numbers in the table are error rates ($\%$). \textbf{M} indicates the distortion magnitude of each transformation. \textbf{RA} for RandAugment~\cite{cubuk2019randaugment}, and \textbf{KD} for knowledege distillation. }
\label{tab:ablation1}
\end{table}


In this modified augment space, we experiment with the strategy of RandAugment~\cite{cubuk2019randaugment} which controls the strength of augmentation by adjusting the distortion magnitude, $M$. For example, on the \texttt{translate-x} transformation, a magnitude of $3$ allows the entire image to be shifted, to the left or right, by at most $30\%$ of the visible field, and a magnitude of $10$ enlarges the number into $100\%$, \textit{i.e.}, the visible area totally disappears. More examples are shown in Figure~\ref{fig:space}. Note that RandAugment performs two consecutive transformations on each image, therefore, a magnitude of $8$ is often enough to destroy all semantic contents. Hence, $M$ is constrained within the range of $0$--$7$ in our experiments. 

Results of different distortion magnitudes are summarized in Table~\ref{tab:ablation1}. With the increase of the magnitude, a larger portion of semantic information is expected to be removed from the training image. In this scenario, if we continue forcing the model to fit the ground-truth, one-hot supervision of each training sample, the deep network may get confused and `under-fit' the training data. This causes consistent accuracy drop, especially in the modified augment space with only geometric transformations. Even when the full augment space is used (in which some transformations are not very sensitive to $M$), this factor persists and hinders the use of larger $M$ values, and thus restricts the degree of freedom of AutoAugment.


Knowledge distillation offers an opportunity that each augmented image is checked beforehand, and a soft label is provided by a pre-trained teacher model to co-supervise the training process so that the deep network is not forced to fit the one-hot label. This is especially useful when the training image is contaminated by augmentation. As shown in Table~\ref{tab:ablation1}, knowledge distillation provides consistent accuracy gain over RandAugment, as it slows down the accuracy drop with aggressive augmentation (the gain is larger as the distortion magnitude increases). More importantly, under a magnitude of ${M}={1}$, knowledge distillation produces an accuracy gain of $1.9\%$, assisting the RandAugment-only model with a deficit of $1.1\%$ to surpass the baseline, claiming an advantage of $0.4\%$. This proves that the benefit mainly comes from the cooperation of RandAugment and knowledge distillation, not only from the auxiliary information provided by knowledge distillation itself~\cite{furlanello2018born,bagherinezhad2018label,yang2019training}.


\begin{table}[!t]
\centering
\setlength{\abovecaptionskip}{10pt}%
\begin{tabular}{p{2.2cm}<{\centering}|p{4cm}<{\centering}|p{0.81cm}<{\centering}p{0.81cm}<{\centering}p{0.81cm}<{\centering}p{0.81cm}<{\centering}p{0.81cm}<{\centering}|p{0.81cm}<{\centering}}
\thickhline
\textbf{Dataset} & \textbf{Network} & \textbf{NA} & \textbf{AA} & \textbf{FAA} & \textbf{PBA} & \textbf{RA} & \textbf{Ours} \\	
\thickhline
{} & Wide-ResNet-28-10 & $3.9$ & $2.6$ & $2.7$ & $2.6$ & $2.7$ & $\textbf{2.4}$ \\
{} &Shake-Shake (26 2x32D) & $3.6$ & $2.5$ & $2.5$ & $2.5$ & $-$ & $\textbf{2.3}$ \\
CIFAR-10 &Shake-Shake (26 2x96D) & $2.9$ & $2.0$ & $2.0$ & $2.0$ & $2.0$ & $\textbf{1.8}$ \\
{} &Shake-Shake (26 2x112D) & $2.8$ & $\textbf{1.9}$ & $\textbf{1.9}$ & $2.0$ & $-$ & $\textbf{1.9}$ \\
{} &PyramidNet+ShakeDrop & $2.7$ & $\textbf{1.5}$ & $1.7$ & $\textbf{1.5}$ & $\textbf{1.5}$ & $\textbf{1.5}$ \\
\hline
{} &Wide-ResNet-28-10 &$18.8$ & $17.1$ & $17.3$ & $16.7$ & $16.7$ & $\textbf{16.2}$ \\
CIFAR-100 & Shake-Shake (26 2x96D) & $17.1$ & $14.3$ & $14.6$ & $15.3$ & $-$ & $\textbf{13.8}$ \\
{} & PyramidNet+ShakeDrop & $14.0$ & $10.7$ & $11.7$ & $10.9$ & $-$ & $\textbf{10.6}$ \\
\thickhline
\end{tabular}
\caption{Comparison between our approach and other data augmentation methods on CIFAR-10 and CIFAR-100. The teacher for all networks is Wide-ResNet-28-10, except for PyramidNet+ShakeDrop with itself as teacher on CIFAR-100 (due to the huge performance gap). All numbers in the table are error rates ($\%$). \textbf{NA} indicates no augmentation is used, \textbf{AA} for AutoAugment~\cite{cubuk2019autoaugment}, \textbf{FAA} for fast AutoAugment~\cite{lim2019fast}, \textbf{PBA} for population-based augmentation~\cite{ho2019population}, and \textbf{RA} for RandAugment~\cite{cubuk2019randaugment}.}
\label{tab:cifar}
\end{table}

\vspace{0.2cm}
\noindent$\bullet$\quad\textbf{Comparison with State-of-the-Arts}

To make fair comparisons to the previous AutoAugment-based methods, we directly inherit the augmentation policies found on CIFAR by AutoAugment. In this full space, all transformations listed in Section~\ref{approach:preliminaries}, not only the geometric transformations, can appear. Results are summarized in Table~\ref{tab:cifar}.

On CIFAR-10, our method outperforms other augmentation methods consistently, in particular, on top of smaller networks (\textit{e.g.}, the error rates of Wide-ResNet-28-10 and two Shake-Shake models are reduced by $0.2\%$). For larger models, in particular PyramidNet with ShakeDrop regularization, the room of improvement on CIFAR-10 is very small, yet we can observe improvement on very large models on the more challenging CIFAR-100 and ImageNet datasets (see the next part for details).

A side comment is that we have used the same teacher model (\textit{i.e.}, Wide-ResNet-28-10, reporting a $2.6\%$ error) which is relatively weak. We find this model can assist training much stronger students (\textit{e.g.}, the Shake-Shake series, in which the error of the 2x96D model, $2.0\%$, is reduced to $1.8\%$). \textbf{In other words, weaker teachers can assist training strong students.} This delivers a complementary opinion to prior research which advocates for extracting `dark knowledge' as some kind of auxiliary supervision~\cite{yang2019training} from stronger~\cite{hinton2015distilling} or at least equally-powerful~\cite{furlanello2018born} teacher models, and further verifies the extra benefits brought by integrating knowledge distillation and AutoAugment together.

On CIFAR-100, we evaluate a similar set of network architectures, \textit{i.e.}, Wide-ResNet-28-10, Shake-Shake (26 2x96D), and PyramidNet+ShakeDrop. As shown in Table~\ref{tab:cifar}, our results consistently outperform the previous state-of-the-arts. For example, on a relatively smaller Wide-ResNet-28-10, the error of AutoAugment decreases from $17.1\%$ to $16.2\%$ and significantly outperforms other methods, \textit{e.g.}, PBA and RA. On Shake-Shake (26 2x96D), our approach also surpasses the previous best performance ($14.3\%$) by a considerable margin of $0.5\%$. On pyramidNet with ShakeDrop, although the baseline accuracy is sufficiently high, knowledge distillation still brings a slight improvement (from $10.7\%$ to $10.6\%$).

\subsection{On the ImageNet Dataset}
\label{experiments:imagenet}

\noindent$\bullet$\quad\textbf{Dataset, Setting, and Implementation Details}

ImageNet~\cite{deng2009imagenet} is one of the largest visual recognition datasets which contains high-resolution images. We use the competition subset which has $1\mathrm{K}$ classes, $1.3\mathrm{M}$ training and $50\mathrm{K}$ validation images. The number of images in each class is approximately the same for training data.

We build our baseline upon EfficientNet~\cite{tan2019efficientnet} and RandAugment~\cite{cubuk2019randaugment}. EfficientNet contains a series of deep networks with different depths, widths and scales (\textit{i.e.}, the spatial resolution at each layer). There are $9$ variants of EfficientNet~\cite{xie2019adversarial}, named from B0 to B8. Equipped with RandAugment, EfficientNet-B7 reports a top-1 accuracy of $85.0\%$ which is close to the state-of-the-art. We start with EfficientNet-B0 to investigate the impact of different knowledge distillation parameters on ImageNet, and finally compete with state-of-the-art results on EfficientNet-B4, EfficientNet-B7, and EfficientNet-B8. 

We follow the implementation details provided by the authors\footnote{\textsf{https://github.com/tensorflow/tpu/tree/master/models/official/efficientnet}}, and reproduce the training process using PyTorch. For EfficientNet-B0, it is trained through $500$ epochs with an initial learning rate to be $0.256$ and decayed by a factor of $0.97$ every $2.4$ epochs. We use the RMSProp optimizer with a decay factor of $0.9$ and a momentum of $0.9$. The batch-normalization decay factor is set to be $0.99$ and the weight decay $10^{-5}$. We use 32 GPUs (NVIDIA Tesla-V100) to train EfficientNet-B0/B4, and 256 GPUs for EfficientNet-B7/B8, respectively. 

\vspace{0.2cm}
\noindent$\bullet$\quad\textbf{The Impact of Different Knowledge Distillation Parameters}

We start with investigating the impact of $\lambda$ and $K$, two important hyper-parameters of knowledge distillation. We perform experiments on EfficientNet-B0 with a moderate distortion magnitude of ${M}={9}$, which, as we have shown in the right-hand side of Figure~\ref{fig:motivation}, is a safe option on EfficientNet-B0. For $\lambda$, we set different values including $0.1$, $0.2$, $0.5$, $1.0$, and $2.0$. For $K$, the optional values include $2$, $5$, $10$, $25$, and $50$. To better evaluate the effect of each parameter, we fix one parameter value when changing the other.

\begin{figure}[!t]
\centering
\includegraphics[width=12cm,height=3.5cm]{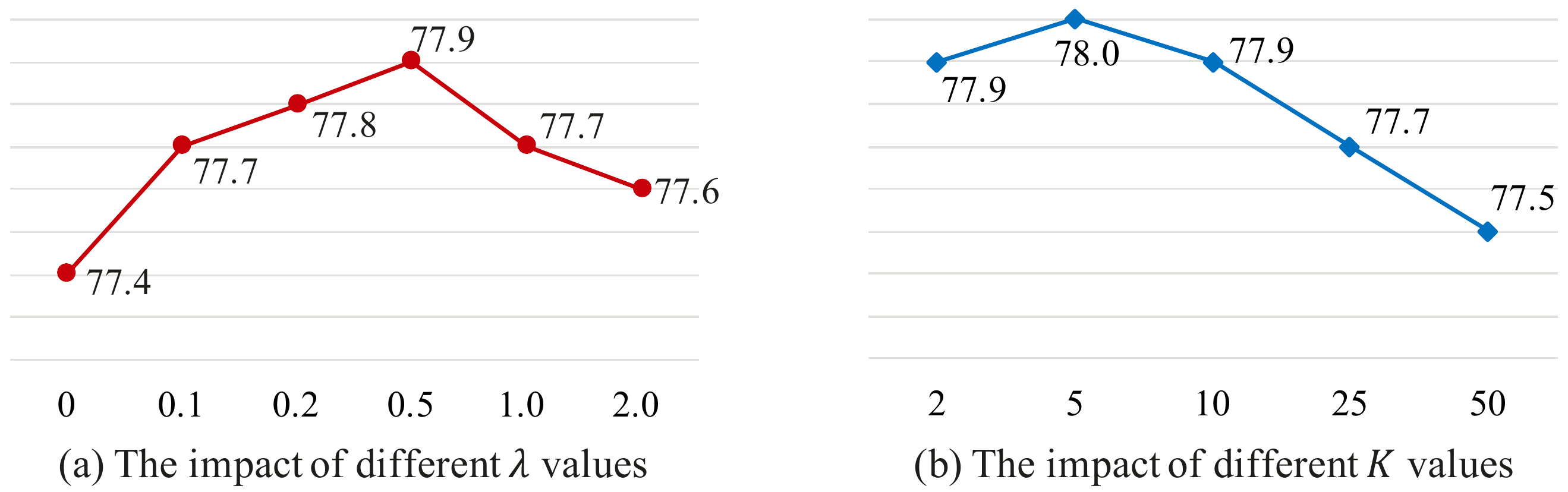}
\caption{Training EfficientNet-B0 with different knowledge distillation parameters. All numbers in the figure are top-1 accuracy ($\%$). \textbf{Left}: The testing accuracy of different $\lambda$ values, while $K$ is set as $10$. \textbf{Right}: The testing accuracy of different $K$ values, while $\lambda$ is set to be $0.5$.}
\label{fig:results_b0_kd}
\end{figure}

Results are shown in Figure~\ref{fig:results_b0_kd}. It is clear that a moderate $\lambda$ performs best. While setting $\lambda$ with a small value, \textit{e.g.}, $0.1$, knowledge distillation is only expected to affect a small part of training samples. Yet, it obtains a $0.3\%$ accuracy gain, implying that these samples, though rarely seen, can make the training process unstable. On the other hand, when $\lambda$ is overly large, \textit{e.g.}, knowledge distillation can dominate the training process and force the student model to have a very similar behavior to the teacher model, which limits its ability and harms classification performance.

Regarding $K$, we note that ${K}={5}$ achieves the best performance, indicating that on average, each class is connected to $4$ other classes. This was also suggested in~\cite{yang2019training}. Yet, we find that setting ${K}={2}$ or ${K}={10}$ reports similar accuracy, but the performance gradually drops as $K$ increases. This implies including too many classes for KL-divergence computation is harmful, because each training image, after augmented with a relatively small distortion magnitude, is not likely to be connected to a large number of classes. However, to train more powerful models, larger distortion magnitudes need to be used and heavier ambiguity introduced. In this case, a larger $K$ will be better, as we shall see in the next section.

Regardless of tuning hyper-parameters, we emphasize that all tested $\lambda$'s, lying in the range of $\left[0.1,2.0\right]$, and all tested $K$'s, in $\left[2,50\right]$, can bring positive effects on classification. This indicates that knowledge distillation is usually useful in training with augmented data. With the best setting, \textit{i.e.}, a distortion magnitude of $9$, a fixed $K$ of $5$, and ${\lambda}={0.5}$, we achieve a top-$1$ accuracy of $78.0\%$ on EfficientNet-B0. This surpasses the accuracy of RandAugment (reproduced by us) and AdvProp~\cite{xie2019adversarial} by margins of $0.6\%$ and $0.4\%$, respectively.

\begin{table}[!b]
\centering
\setlength{\abovecaptionskip}{10pt}%
\begin{tabular}{p{3cm}<{\centering}|p{3cm}<{\centering}|p{0.95cm}<{\centering}p{0.95cm}<{\centering}p{0.95cm}<{\centering}p{1.55cm}<{\centering}|p{0.9cm}<{\centering}}
\thickhline \textbf{Teacher Network} & \textbf{Student Network}  & \textbf{AA} & \textbf{RA} & \textbf{RA$^{\dag}$} & \textbf{AdvProp} & \textbf{Ours} \\
\thickhline
 EfficientNet-B0 & EfficientNet-B0 & $77.3$ & $-$ & $77.4$ & $77.6$ & \textbf{78.0} \\
 EfficientNet-B4 & EfficientNet-B4 & $83.0$ & $-$ & $83.0$ & $83.3$ & \textbf{83.6} \\
 EfficientNet-B7 & EfficientNet-B7 & $84.5$ & $85.0$ & $84.9$ & $85.2$ & \textbf{85.5} \\
EfficientNet-B7* & EfficientNet-B8 & $84.8$ & $85.4$ & $-$ & $85.5$ & \textbf{85.7} \\
\thickhline
\end{tabular}
\vspace{0.1cm}
\caption{Comparison between our approach and other data augmentation methods on ImageNet. All numbers in the table are top-$1$ accuracy ($\%$). \textbf{AA} indicates AutoAugment~\cite{cubuk2019autoaugment} is used, \textbf{RA} for RandAugment~\cite{cubuk2019randaugment}, and \textbf{AdvProp} for Adversarial Propagation method~\cite{xie2019adversarial}. \textbf{RA$^{\dag}$} denotes the results of RandAugment produced by ourselves in PyTorch. EfficientNet-B7* denotes the student model in the penultimate row, which achieves a top-$1$ accuracy of $85.5\%$.}
\label{tab:results_imagenet}
\end{table}

\vspace{0.2cm}
\noindent$\bullet$\quad\textbf{Comparison to the State-of-the-Arts}


To better evaluate the effectiveness of our approach, we further conduct experiments on more challenging large models,~\textit{i.e.}, EfficientNet-B4, EfficientNet-B7, and EfficientNet-B8. Given the fact that larger network is expected to over-fit more easily, for EfficienNet-B4 and EfficientNet-B7, we lift the magnitude of transformations on RandAugment from $9$ in EfficientNet-B0 to $15$ and $28$, respectively. As discussed above, increasing the distortion magnitude brings more ambiguity to the training images so that each of them should be connected to more classes, and the knowledge distillation supervision should take a heavier weight. Hence, we increase $K$ to $50$ and $\lambda$ to $2.0$ in all experiments in this part.

Results are summarized in Table~\ref{tab:results_imagenet}. By restraining the inevitable noises generated by RandAugment, our approach significantly boosts the baseline models. As shown in Table~\ref{tab:results_imagenet}, the top-1 accuracy of EfficientNet-B4 is increased from $83.0\%$ to $83.6\%$, and that of EfficientNet-B7 from $84.9\%$ to $85.5\%$. The margin of $0.6\%$ is considered significant in such powerful baselines. Both numbers surpass the current best, AdvProp~\cite{xie2019adversarial}, without using adversarial examples to assist training.

Following AdvProp~\cite{xie2019adversarial}, we also move towards training EfficientNet-B8. The hyper-parameters remain the same as in training EfficientNet-B7. Due to GPU memory limit, we use the best trained EfficientNet-B7 (with a $85.5\%$ accuracy) as the teacher model. We report a top-1 accuracy of $85.7\%$, which sets the \textbf{new state-of-the-art} on the ImageNet dataset (without extra training data). With the test image size increased from $672$ to $800$, the accuracy is slightly improved to $85.8\%$. We show the comparison with previous best models in Table~\ref{tab:SOTA}.

\begin{table}[!t]
\centering
\setlength{\abovecaptionskip}{10pt}%
\begin{tabular}{c|c|c|c}
\thickhline
\textbf{Method} & \textbf{Params} & \textbf{Extra Training Data} & \textbf{Top-1 ($\%$)} \\
\thickhline
ResNet-152~\cite{he2016deep} & $60$M & $-$ & $77.8$ \\
DenseNet-264~\cite{huang2017densely} & $34$M & $-$ & $77.9$ \\
Inception-v4~\cite{Szegedy2017Inception} & $48$M & $-$ & $80.0$ \\
ResNeXt-101~\cite{Xie2017Aggregated} & $84$M & $-$ & $80.9$ \\
SENet~\cite{Hu2018Squeeze} & $146$M & $-$ & $82.7$ \\
NASNet-A~\cite{zoph2018learning} & $89$M & $-$ & $82.7$ \\
AmoebaNet-C~\cite{Real2019Regularized} & $155$M & $-$ & $83.5$ \\
GPipe~\cite{Huang2019Gpipe} & $557$M & $-$ & $84.3$ \\
EfficientNet-B7~\cite{cubuk2019randaugment} & $66$M & $-$ & $85.0$ \\
EfficientNet-B8~\cite{cubuk2019randaugment} & $88$M & $-$ & $85.4$ \\
EfficientNet-L2~\cite{tan2019efficientnet} & $480$M & $-$ & $85.5$ \\
AdvProp (EfficientNet-B8)~\cite{xie2019adversarial} & $88$M & $-$ & $85.5$ \\
\hline
{ResNeXt-101 Billion-scale~\cite{Yalniz2019Billion}} & {$193$M} & {3.5B tagged images} & {$84.8$} \\
{FixRes ResNeXt-101 WSL~\cite{Touvron2019Fixing}} & {$829$M} & {3.5B tagged images} & {$86.4$} \\
\textcolor{blue}{Noisy Student (EfficientNet-L2)~\cite{xie2019self}} & \textcolor{blue}{$480$M} & \textcolor{blue}{300M unlabeled images} & \textcolor{blue}{$88.4$} \\
\hline
{Ours (EfficientNet-B7 w/ KD)} & {$66$M} & {$-$} & {$85.5$} \\
\textcolor{red}{Ours (EfficientNet-B8 w/ KD)} & \textcolor{red}{$88$M} & \textcolor{red}{$-$} & \textcolor{red}{$85.8$} \\
\thickhline
\end{tabular}
\caption{Comparison to the state-of-the-arts on ImageNet. In the middie panel, we list three approaches with extra training data (a large number of weakly tagged or unlabeled images). \textcolor{red}{Red} and \textcolor{blue}{blue} texts highlight the best results to date without and with extra training data, respectively.}
\label{tab:SOTA}
\end{table}

\section{Conclusions}
\label{conclusions}

This paper integrates knowledge distillation into AutoAugment-based methods, and shows that the noises introduced by aggressive data augmentation policies can be largely alleviated by referring to a pre-trained teacher model. We adjust the computation of KL-divergence, so that the teacher and student models share similar probabilistic distributions over the top-ranked classes. Experiments show that our approach indeed suppresses noises introduced by data augmentation, and thus stabilizes the training process and enables more aggressive AutoAugment policies to be used. On top of EfficientNet and RandAugment, we set the new state-of-the-art, a $85.8\%$ top-1 accuracy, on the ImageNet dataset (without extra training data).

In spite of the consistent improvement brought by our approach, there are still problems that remain mostly uncovered. For example, it remains unclear if useful information only exists in top-ranked classes determined by the teacher model, and whether mimicking the class-level distribution is the optimal choice. We will continue investigating these topics in our future research.

\vspace{0.2cm}
\textbf{Acknowledgements}\quad We thank Jianzhong He for helping with setting up the parallelized training system. We thank Chunjing Xu, Wei Zhang, and Zhaowei Luo for coordinate hardware resource. We also thank Zhengsu Chen, Yuhui Xu, Lanfei Wang, and Kaifeng Bi for instructive discussions.

\bibliographystyle{splncs04}
\bibliography{egbib}
\end{document}